\documentclass{article}

\usepackage{epsfig, graphics}
\usepackage{latexsym,subfigure}
\usepackage{amsmath,amssymb,amsthm,amsfonts}
\usepackage[numbers]{natbib}
\usepackage{algorithm,algorithmic,bm, color}

\title{Scalable Audience Reach Estimation in Real-time Online Advertising}

\author{Ali Jalali\\ajalali@turn.com\\Turn Inc\\
\and
Santanu Kolay\\skolay@turn.com\\Turn Inc\\
\and
Peter Foldes\\pfoldes@turn.com\\Turn Inc\\
\and
Ali Dasdan\\adasdan@turn.com\\Turn Inc}

\begin{document}
\maketitle

\begin{abstract}
Online advertising has been introduced as one of the most efficient methods of advertising throughout the recent years. Yet, advertisers are concerned about the efficiency of their online advertising campaigns and consequently, would like to restrict their ad impressions to certain websites and/or certain groups of audience. These restrictions, known as targeting criteria, limit the reachability for better performance. This trade-off between reachability and performance illustrates a need for a forecasting system that can quickly predict/estimate (with good accuracy) this trade-off. Designing such a system is challenging due to (a) the huge amount of data to process, and, (b) the need for fast and accurate estimates. In this paper, we propose a distributed fault tolerant system that can generate such estimates fast with good accuracy. The main idea is to keep a small representative sample in memory across multiple machines and formulate the forecasting problem as queries against the sample. The key challenge is to find the best strata across the past data, perform multivariate stratified sampling while ensuring fuzzy fall-back to cover the small minorities. Our results show a significant improvement over the uniform and simple stratified sampling strategies which are currently widely used in the industry. 
\end{abstract}

\section{Introduction}
\label{sc:introduction}

Forecasting is the problem of predicting the future of a time series considering the past. When it comes to big data, generating a fast and accurate report of the past is a challenge itself. Typically advertisers would like to frequently forecast on thousands of dimensions to have a good understanding of the market and spend their budget wisely. Running such queries against massive data takes a lot of resources and time making forecasting impractical. In practice, it is expected to return the forecasting result in few seconds so that the advertisers can adjust their audience and contextual targeting.

One way of generating fast and accurate reports is to take a relatively small sample from the original big data and run queries against that sample. At the end, the result can be scaled up to compensate for the sampling rate. In such a scheme, the accuracy of the report depends on how good a representative the sample is for the entire big dataset. Since the data lies in a high dimensional space, with thousands of dimensions, it is challenging to get a good representative of the data. In particular, uniform sampling performs poorly in this regime, despite its popular use in the industry.

In this paper, we cast the problem of generating fast and accurate reports as a sampling problem and introduce metrics to measure the goodness of the report. The sampling algorithm should be able to guarantee a good accuracy while it is practically feasible on big data. Also, since the samples are going to be used in a distributed system, the sampling algorithm should be able to produce reasonable results even if some parts of the system fail, i.e., some portion of the samples are not available at a certain point. We propose both a robust and accurate sampling algorithm as well as the architecture of the distributed system that uses those samples to generate the results. 

The proposed sampling algorithm is a stratified sampling algorithm that has four steps. In the first step, the algorithm identifies the right set of strata in the high-dimensional data space to be sampled from. Second step involves with a fuzzy fall-back step that maps minor strata to (related) major strata to ensure the diversity and accuracy. The distributed stratified sampling itself is implemented in the third step. The last step is to distribute the samples across multiple machines to make report generation fast and the system fault tolerant. We explain each of these steps in separate \S~\ref{sc:strata}-\ref{sc:distributed}. Experimental results and discussion is provided in \S~\ref{sc:exp}; showing the effectiveness of our method in real application. The paper is finally concluded in section \S~\ref{sc:conclusion}.

\begin{figure*}[t]
\centering
\subfigure[Original Data]{
    \includegraphics[width=2.2in]{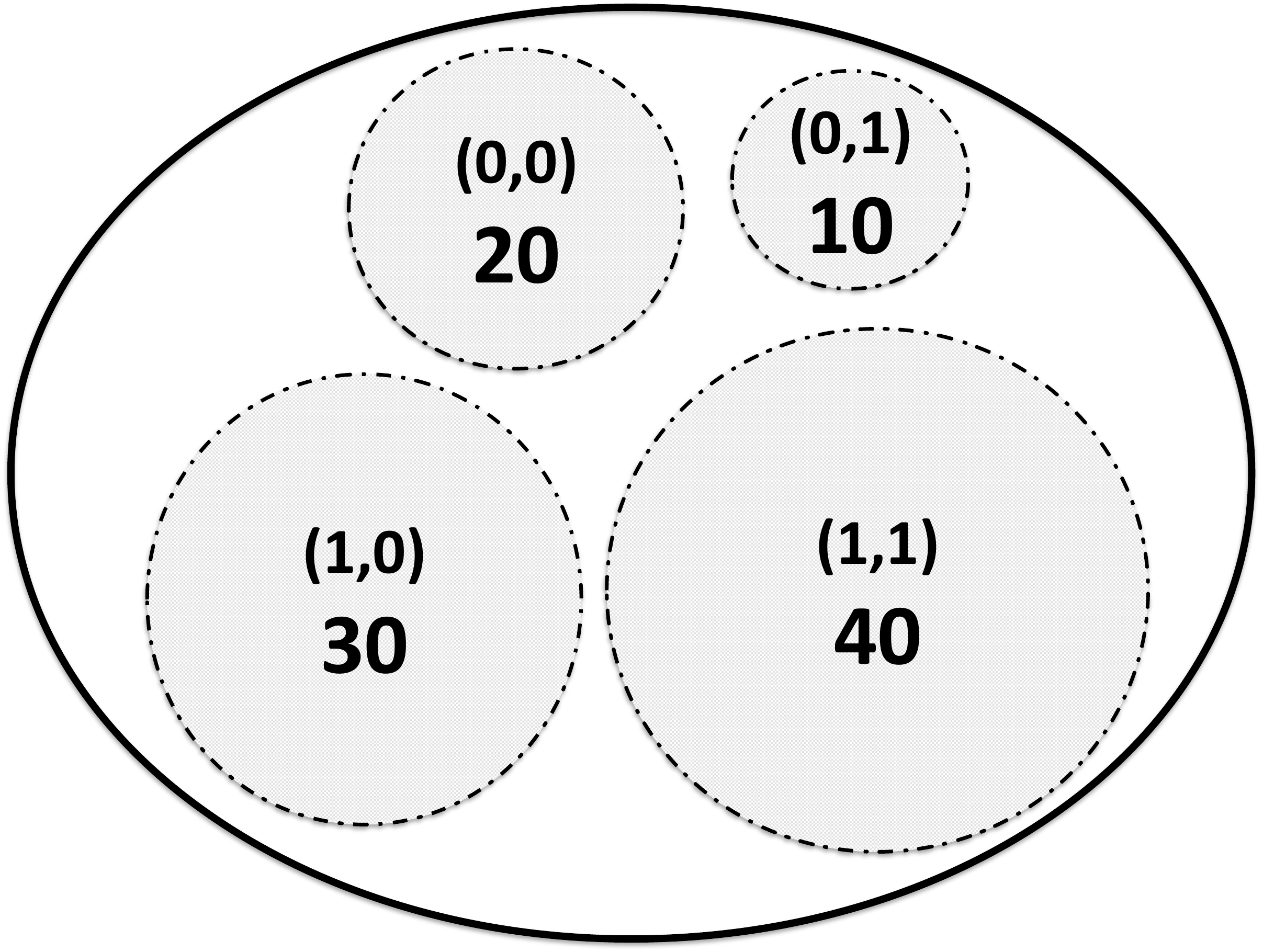}
    \label{fig:original}
}
\subfigure[Accurate Sample]{
    \includegraphics[width=2.2in]{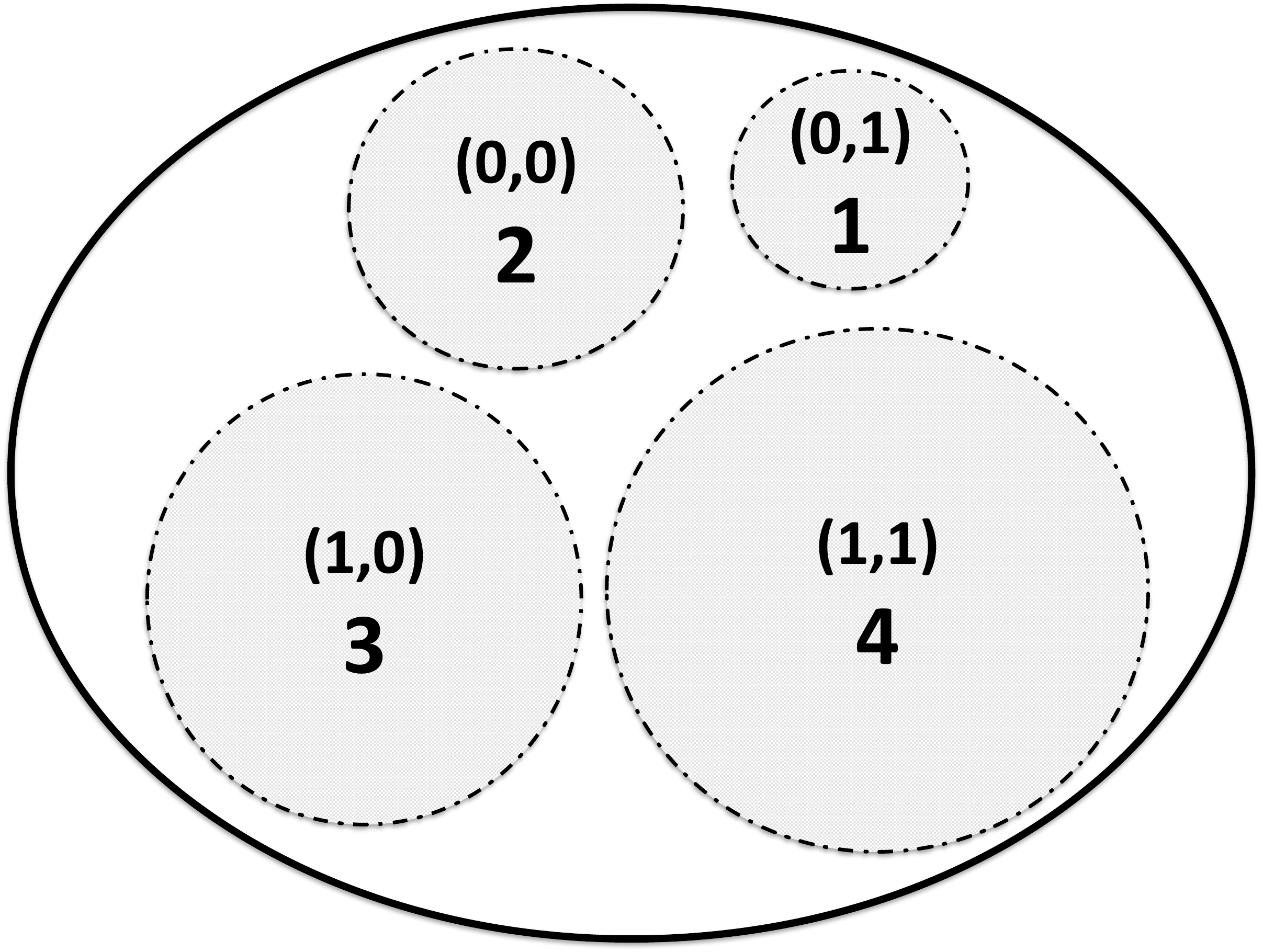}
    \label{fig:correct}
}
\subfigure[Inaccurate Sample]{
    \includegraphics[width=2.2in]{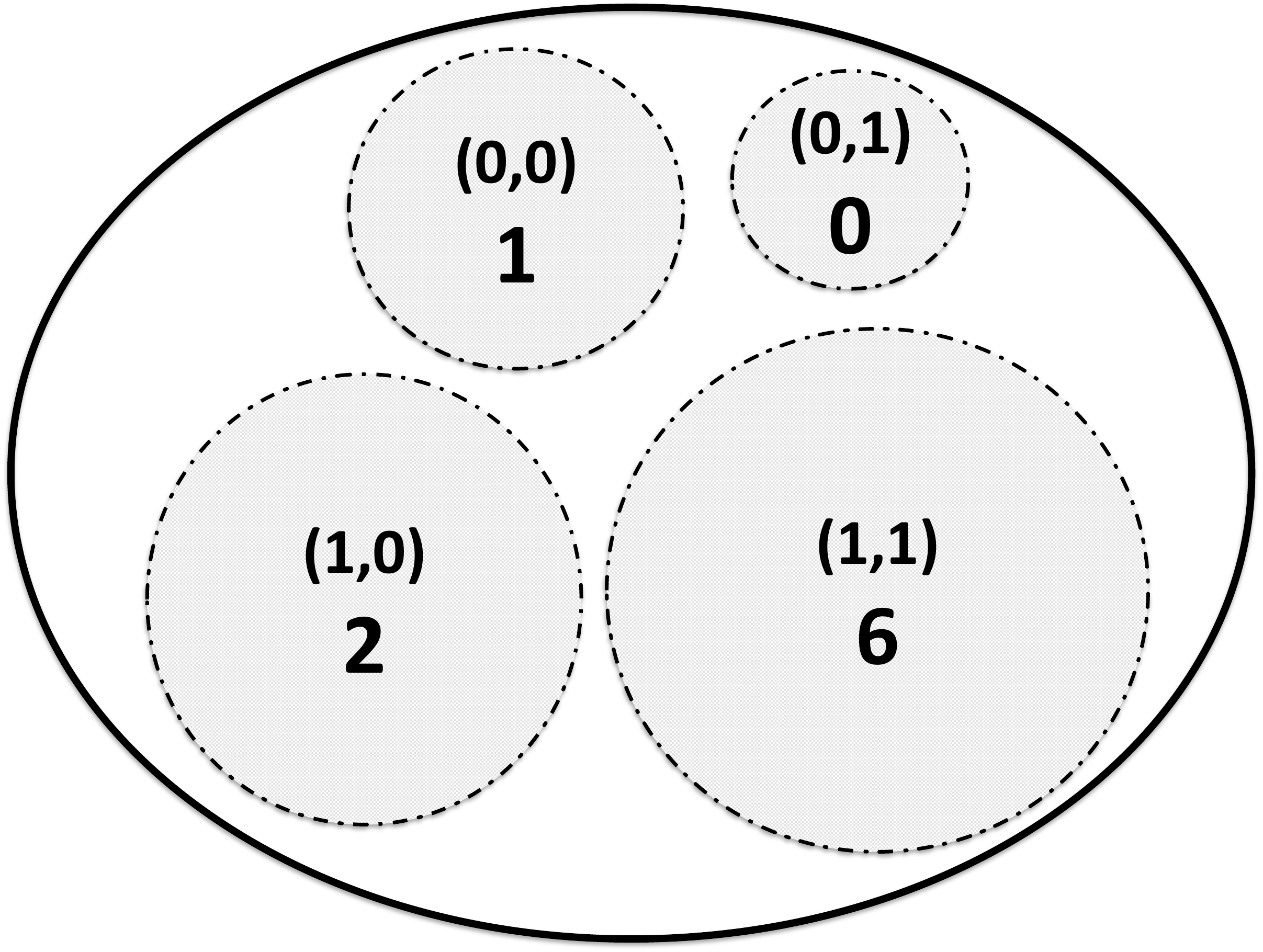}
    \label{fig:incorrect}
}
\caption{A simple toy example to illustrate the problem of uniform sampling.}
\label{fig:ToyExample}
\end{figure*}

\section{Background and Related Work}
\label{sc:background}

In this section, we first formalize the problem and explain the setup. Then, we explore the related work in the context of our problem. We cannot cover all of the literature; rather, we cite more recent works that are closer to our setting and formulation.

\subsection{Problem Setup and Motivation}

Suppose our big data is in the form of a set $\mathcal{D}\!=\!\{\!\mathbb{X}_1,\!\ldots\!,\mathbb{X}_N\!\}$ of $K$-feature vectors $\mathbb{X}_i=(x_1,\ldots,x_K)$ for all $i\in\{1,\ldots,N\}$. Practically, we can assume all features are categorical and if not, we bucketize them. Without loss of generality, assume that each feature $x_j$ can take $m_j$ different values from $\{1,\ldots,m_j\}$. We would like to find a good representative sample set $\mathcal{D}_s\subset\mathcal{D}$ such that $n=|\mathcal{D}_s|<\!\!<N$, i.e., the size of the sample is much smaller than the size of the original dataset. We need to define a metric to compare different sample sets and prefer one to the other in a consistent way.

Consider the set of all queries $\mathcal{Q}$ that want to count the number of $K$-feature vectors lie in a certain subspace. The cardinality of this set is $2^{\prod m_j}$ since there are $\prod_{j=1}^K m_j$ different possibilities for each feature vector. Suppose we submit a queries $q_i\in\mathcal{Q}$ to both $\mathcal{D}$ and $\mathcal{D}_s$ and we get $N_i$ and $n_i$, respectively. We expect that $N_i\approx\frac{N}{n}n_i$ and hence, we can define the following error metric:
\begin{equation}
\text{\bf err}(\mathcal{D}_s) = \max_{q_i\in\mathcal{Q}}\; \frac{\left|N_i - \frac{N}{n}n_i\right|}{N_i} =\max_{q_i\in\mathcal{Q}}\; \left|1 - \frac{N}{n}\frac{n_i}{N_i}\right|.
\end{equation}
The smaller $\text{\bf err}(\mathcal{D}_s)$ is, the better representative $\mathcal{D}_s$ is for our big data. Notice that although normalized on the true size of the query $N_i$, this error can exceed $1$ simply if the over estimation is too large.

Before we proceed, it is essential to understand why uniform sampling cannot be a solution in this high-dimensional regime. Imagine we build $\mathcal{D}_s$ by uniformly (with probability $\frac{n}{N}$) sampling feature vectors from $\mathcal{D}$. For all possible query $q_i$ in the high-dimensional space, we require $\frac{n_i}{N_i}\approx\frac{n}{N}$. It is easy to see that satisfying so many requirements at the same time is highly unlikely. To show this, we provide a toy example.

Suppose our big data has $N=100$ binary feature vectors and $K=2$, i.e., each feature vector takes a value in $\{(0,0),(0,1),(1,0),(1,1)\}$. The population of each of these vector values is shown in Fig.~\ref{fig:original}, e.g., we have 20 feature vectors equal to $(0,0)$ in our data. We would like to take a representative sample with $n=10$. Fig.~\ref{fig:correct} illustrates a good representative sample $\mathcal{D}_s$ of the data, i.e., it picks two feature vectors from $(0,0)$, one feature vector from $(0,1)$, etc. For any query on this $\mathcal{D}_s$, if we multiply the result $n_i$ by $\frac{N}{n}=10$, we get the true result of that query $N_i$ as if we were running that query on the original data. In this case, it is easy to see that $\text{\bf err}(\mathcal{D}_s)=0$. However, with uniform sampling, we are not guaranteed to get such a nice sample even though (like this case), our sample size is big enough. With uniform sampling, we might end up with a sample set that looks like Fig.~\ref{fig:incorrect}.

It is natural to ask with what probability the uniform sampling produces the minimum error sample. In the case of our toy example (see Fig.~\ref{fig:ToyExample}), this probability can be expressed as
\begin{equation}
P = \frac{{10 \choose 1}{20 \choose 2}{30 \choose 3}{40 \choose 4}}{{100 \choose 10}}\approx 0.04.
\end{equation}
This probability is shockingly low and it becomes worse as the number of values (aka strata) that the feature vector can take increases. This increase is inevitable in high dimensions. More generally, if we have $M$ strata, i.e., feature vectors take $M$ different (vector) values, with sizes $N_1,\ldots,N_M$ with $\sum_i N_i = N$, and, our sampling size is $n$, the probability that uniform sampling achieves the lowest error has a hyper-geometric distribution as
\begin{equation}
P^* = \frac{\left(\begin{aligned}&\text{ }\quad N_1\\ &\left[\frac{n}{N}N_1\right]\end{aligned}\right)\left(\begin{aligned}&\text{ }\quad N_2\\ &\left[\frac{n}{N}N_2\right]\end{aligned}\right)\ldots\left(\begin{aligned}&\text{ }\quad N_M\\ &\left[\frac{n}{N}N_M\right]\end{aligned}\right)}{\left(\begin{aligned}&N\\ &n\end{aligned}\right)}.
\end{equation}
Using the Normal approximation summarized in Theorem~2 of \cite{nicholson1956normal}, we have
\begin{equation}
P^*\leq\frac{N!\left(\frac{e}{n}\right)^n}{\left(\sqrt{2\pi}\right)^M(n-M+1)}.
\label{eq:uniform-err}
\end{equation}
This probability vanishes as $M$ increases; implying that uniform sampling cannot produce a good representative even with large enough sample size. This motivates consideration of stratified sampling; however, implementing stratified sampling in high dimensions is a challenge. Further, we will discuss that stratified sampling alone is not enough and introduce a fuzzy fall-back step.

\subsection{Related Work}

There is large body of work in literature related to Approximate Query Processing (AQP). There are three main approaches to AQP-Histogram based \cite{acharya2000congressional, bruno2001stholes, gunopulos2000approximating, cuzzocrea2009lcs}, wavelet based \cite{chakrabarti2001approximate, matias1998wavelet} and sampling based \cite{gibbons1998new, jermaine2008scalable}. Garofalakis \emph{et al.} \cite{garofalakis2001approximate} provides a tutorial introduction to the subject. Our work is most closely related to the sampling based and histogram based methods.

The work whose approach is closest to our method is by Acharya \emph{et al.} \cite{acharya2000congressional}. They solve the problem of approximating GROUP BY queries using a smaller sample. They propose to use a mixture of uniform (House) and biased (Senate) sampling techniques. For groups with high selectivity the uniform sampling produces the best result, whereas for smaller groups biased (stratified sampling) produces better results. They also propose an algorithm (Congressional sampling) to mix the sampling approach to produce a fixed size sample that provides good error bounds for all groups under the expected query load. In their biased sampling approach, they use all possible subsets of the grouping columns to find the strata, each strata basically represents one possible row of a GROUP BY query. However, it is not clear how this algorithm scales for tables with very large number of columns, as is the case in our problem. Another main drawback of this paper is the assumption about highly selective queries: They ignore cases where samples generated out of congressional sampling may not contain any row where GROUP BY keys has very log selectivity. In our approach we must account for it; otherwise, we are not able to forecast for some types of constraints entered by our users leading to bad user experience. Also many of our forecast related queries especially geographical based constraints are generally very selective in nature.

Chaudhuri \emph{et. al.} \cite{chaudhuri2007optimized} provide a stratified sampling based approach to approximate query processing under a workload distribution. The authors propose the STRAT algorithm, which first defines a set of strata called ``Fundamental Regions''. These fundamental regions are chosen based on an expected query load with the constraint that whenever a fundamental region is used all elements from that region must be selected. The authors show good result with this approach for SUM and COUNT type of queries based on GROUP BY constraints. However this approach does not fit our use case exactly, as we do not have an estimate of the expected query load beforehand. Also, the paper does not clearly mention the sample generation complexity for high dimensional dataset. However the authors provide a very good analysis of variance of different types of queries and show their method performs quite well under diverse set of query workload.

Authors of \cite{ganti2000icicles} introduce a novel way of maintaining dynamic samples for AQP. They use ``icicles'', a type of self-tuning samples that produces smaller errors for more frequent query types. For each query they initially generate the result out of full dataset and then choose a uniform sample from the result set. As new queries arrive, the sample is augmented with other rows matching this query. Rows selected multiple times across multiple queries keeps accumulating higher weight. Over time the sample contains a subset of rows from the original dataset with each selected row having a weight proportion to the no of queries that row answers. The authors show that icicles provide significantly better results than uniform sampling; however, it is not clear how the method can control the overall sample size, as in our implementation we have limited space available across all strata and we are expected to provide reasonable accuracy on a set of unforeseen queries.

In \cite{cuzzocrea2009lcs}, authors study the problem of building multidimensional histograms to answers queries approximately over very high dimensional data cubes.  They propose LCS-Hist, which processes the data offline and builds multidimensional histogram to be used as pre-aggregates during queries later. Their method first partitions the $n$-dimensional cube into a set of partitions with the objective of minimizing within partition variance using a dynamic programming approach. Then they try to merge buckets having similar distribution based some distance metric. After several types of bucket merging, the number of buckets become lot more manageable. The authors report significant space reduction by maintaining the aggregates using the histograms compared to the full cube. Even though the range splitting and bucket merging has some similarity with our approach, our fundamental problem is to derive a set of user samples that can be used for arbitrary queries, not necessarily OLAP style \cite{li2004high} aggregates. Even though the authors provide very detailed analysis of approximation error compared to other histogram based methods, they do not provide any result on the computation complexity of their method. As our system has to generate a sample of millions of rows from an original data set of billions of rows, the runtime of our algorithm is of paramount importance.

Some authors have proposed probabilistic methods for selectivity estimation \cite{getoor2001selectivity}.  They model the data set as joint distribution over the variables on which GROUP BY or JOIN queries are run. A naive approach to modeling the joint probability distribution will lead to exponential no of entries, so they propose to use conditional independence found in many real-life data sets. Based on the input query and the conditional independence among the various variables, a Bayesian Network is formulated and answering the query simply boils down to inference on the Bayesian Network. Our approach has some similarity to thier work in the sense that we also rely on the conditional independence property to reduce  our computation.  However the authors seem to focus more on the model complexity and do not provide much informaation on the runtime of the model building and inference. We propose to use a state of the art Markov Random Field method which is learned in distributed environment.

The problem of estimating the result of aggregation query with low selectivity has been tackled by other work related to deep Web \cite{liu2012stratified}, with the additional constraint of not having full access to the data set. This method proposes a stratification method based on a queriable auxiliary attribute based on which the breaking points in the strata are determined. The breaking points in the strata are computed based on a novel Bayesian Adaptive Harmony Search algorithm. The problem solves the sample allocation problem as a constrained optimization problem where the optimization metric consists of both sample variance as well as precision.

To the best of our knowledge, our problem setting is quite unique. Though our method bears some resemblance to the methods proposed in \cite{acharya2000congressional, cuzzocrea2009lcs}, our data set is very high-dimensional (250K) and consists of categorical variables only. This allows us to pose the problem in terms of sets of objects and we can perform stratification based on set operations.

\section{Identifying Strata}
\label{sc:strata}
The first step towards a better sampling is to identify subsets of similar feature vectors, known as \emph{strata}. In high-dimensions, it is a challenge to partition the data into strata such that each stratum is not too big to include non-similar feature vectors and not too small to require a large sample size. In this section, we propose a method for identification of such strata.

\begin{figure}[t]
\centering
\includegraphics[width=2.0in]{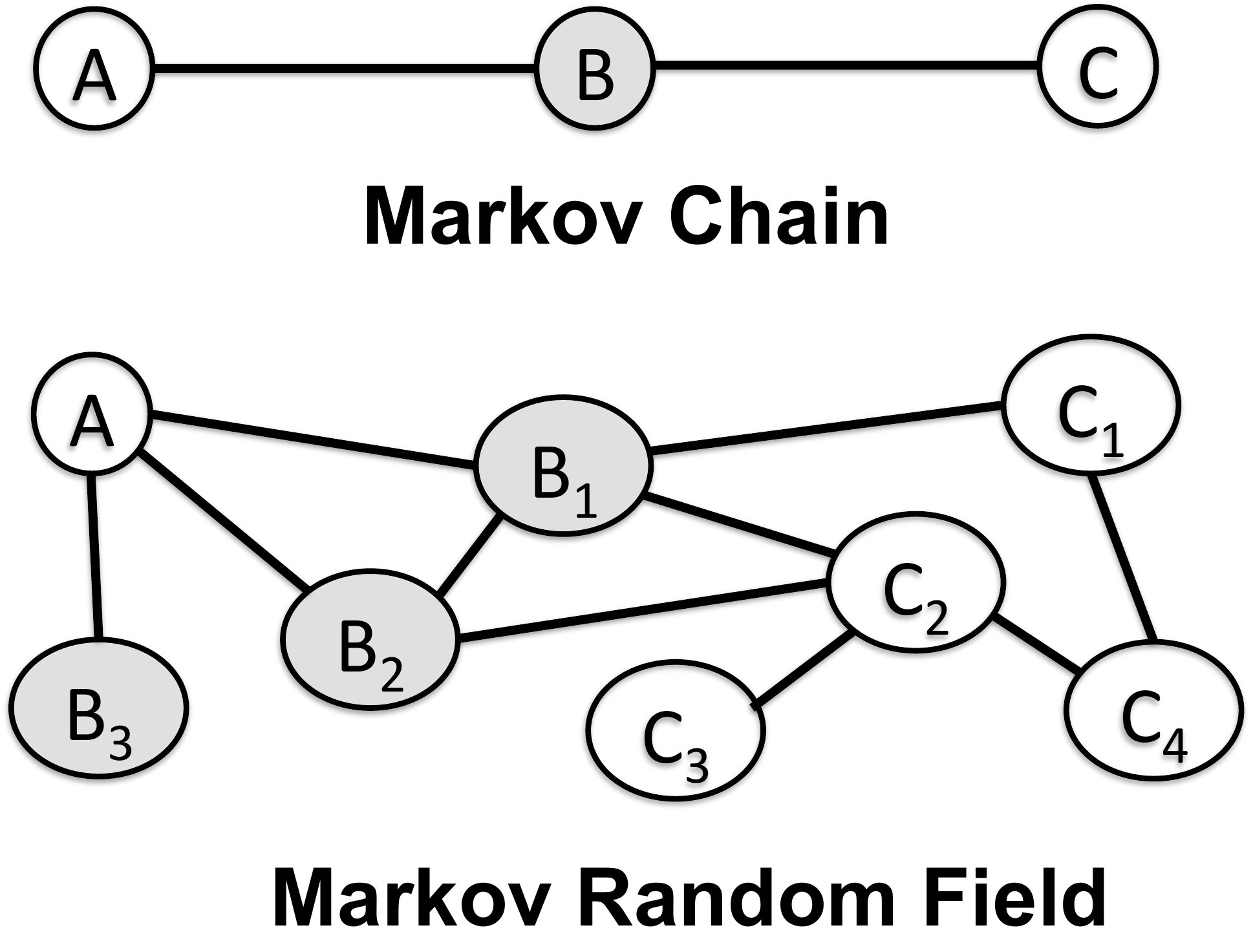}
\caption{\small An illustration of the similarity between Markov chain and Markov Random Field. In the Markov chain, $A$ is independent of $C$ given $B$; similarly, in the Markov random field, $A$ is independent of $(C_1,C_2,C_3,C_4)$ given $(B_1,B_2,B_3)$.}
\label{fig:MRF-Definition}
\end{figure}

Although the feature vectors live in a high-dimensional space, not all of the features are independent. This is the key to dimensionality reduction. Suppose we find a subset of features, i.e., dimensions, that every other feature is highly correlated with this sets. Then, if we build our strata by reducing the dimension of the feature vectors to that subset, we can represent the entire population fairly well. To find such subset of features we use Markov Random Fields (MRFs) \cite{kindermann1980markov}. 

Considering each feature as a node in the graph, MRF assigns a matrix $\Theta_{j,k}\in\mathbb{R}^{m_j\times m_k}$ to the edge intersecting features $x_j$ and $x_k$ such that
\begin{equation}
\begin{aligned}
\mathbb{P}\big[x_j = t \big|\; x_1,\ldots&,x_{j\!-\!1},x_{j\!+\!1},\ldots,x_K\big]\\ &\propto \exp\left(-\sum_{k\neq j}\sum_{q=1}^{m_k}\Theta_{j,k}(t,q)\mathbf{1}_{\{x_k = q\}}\right)
\end{aligned}
\label{eq:MRF-Prob}
\end{equation}
where, $\Theta_{j,k}(t,q)$ is the $(t,q)$ element of the matrix $\Theta_{j,k}$ and $\mathbf{1}_{\{\cdot\}}$ is the indicator function. If we remove the edges with matrix $\Theta_{j,k}$ equal to zero, the remaining graph has the property that each feature conditioned on its neighbors in the graph is independent from the rest of the graph \cite{kindermann1980markov}. See Fig.~\ref{fig:MRF-Definition}. This is a generalization of the Markov Chain where each event given the immediate previous event is independent from the history of events. A distributed and fast algorithm to learn the MRF graph from data is detailed in \cite{jalalilearning}.

Once the MRF graph among features is learned, we need to find a minimum vertex cover \cite{cormen2001introduction} of that graph. Basically, we need to find a minimum subset of vertices such that if we remove them, the rest of the graph becomes completely disconnected. Then, according to MRFs, those vertices are independent of each other given the vertices in our subset. As an example, suppose in the MRF shown in Fig.~\ref{fig:MRF-Definition}, if we select features $\{A,B_2,C_1,C_2\}$ as our features, the rest of the graph become conditionally independent.

Minimum vertex cover is known to be an NP-hard problem; however, there are good approximations such as the algorithm proposed in \cite{chen2001vertex,papadimitriou1998combinatorial}. Without loss of generality, suppose we select features $\{1,\ldots,K^*\}$ (out of $\{1,\ldots,K\}$). For each value that each feature can take, we generate a key-value tag, e.g., ``$x_1=1$" is a tag. There are the total of $\prod_{j=1}^{K^*}m_j$ different tags. Now, each feature vector can be tagged with the set of relevant tags, e.g, the feature vector $(\alpha_1,\alpha_2,\ldots,\alpha_K)$ will be tagged (only on the first $K^*$ features) by the set of tags $\{$``$x_1=\alpha_1$"$,\ldots,$``$x_{K^*}=\alpha_{K^*}$"$\}$.

From a different point of view, each tag can be seen as a set and each feature vector lies in the intersection of the sets corresponding to its tags. This view represents a partitioning on the feature space. We pick each of those partitions as a stratum. Another advantage of this tagging strategy is that if we have one (or some) missing data, we will only lose a tag and the feature vector will be moved to the intersection of the relevant $K^*-1$ sets. The only issue here is that the strata might be very small to sample from and hence, we introduce a fuzzy fall-back step in the next section.

\section{Fuzzy Fall-back}
\label{sc:fuzzy}

Consider the strata we found in the previous sections in the form of intersection of sets. Let $\mathcal{T}_v$ is the set of all feature vectors that are tagged with the tag $v$. As defined before, a tag $v$ is of the form ``$x_i=m$" for the $i^{th}$ feature taking value $m$. Considering strata as the partitions defined by all possible intersections of sets $\mathcal{T}_v$, some of these strata include more than $\frac{N}{2n}$ feature vectors and hence, they can potentially have at least one representative in our stratified sample. However, in reality, most of the such strata do not have enough population. That is why we introduce a fall-back step.

Let $\Upsilon_1$ be the set of all strata with more than $\frac{N}{2n}$ in population size and $\Upsilon_2$ to be the set of all strata that only belong to one tag set; e.g. $\mathcal{T}_v - \bigcup_{u\neq v}\mathcal{T}_u$ is an example of such strata; and, finally let $\Upsilon = \Upsilon_1 \cup \Upsilon_2$ be the set of \emph{stable} strata. Next, we are going to map \emph{unstable} strata $\Upsilon^\perp$ into $\Upsilon$ so that the chance of having small population strata decreases.

We propose two algorithm for the mapping of unstable strata to stable ones. We call this step as fuzzy fall-back step since we change the membership of feature vectors across strata by falling back into their subsets. Our first proposed algorithm has lower computational complexity but slightly more inaccurate. The second algorithm provides a better stratification results at the cost of some computational cost.
 
\begin{itemize}
\item {\bf Algorithm 1}: For each unstable stratum $\mathcal{P}\in\Upsilon^\perp$, find the set of all stable strata $\mathcal{Q}_1,\ldots,\mathcal{Q}_{r}\in\Upsilon$ that are a subset of $\mathcal{P}$ with maximum number of tags. Among $\mathcal{Q}_i$'s, pick $\mathcal{Q}$ that has the smallest population and consider the union of stratum $\mathcal{P}$ and $\mathcal{Q}$ as a new stratum.

\item {\bf Algorithm 2}: This algorithm, at each step, starts with an unstable stratum $\mathcal{P}$ that is in the intersection of maximum number of sets. Then, it distributes the population of $\mathcal{P}$ among the subset strata (regardless whether they are stable or unstable) proportional to their size. After distribution, some of those subsets might become stable, i.e., get more than $\frac{N}{2n}$ in population size and hence get removed from the set of unstable strata. The process continues until there is no unstable strata or there is no unstable strata with a subset.
\end{itemize}

After this step, we can simply sample from each strata according to their size. Notice that with this strategy, if $\mathcal{D}_s$ is our sample, for each strata, we have
\begin{equation}
\text{\bf err}(\mathcal{D}_s) \leq\frac{N}{2n}.
\end{equation}
This error bound is far better than the uniform bound shown in \eqref{eq:uniform-err}. The only caveat is that when we want to make estimation in real system, we have queries against all features not just the ones we selected in Section~\ref{sc:strata}. To compensate for that, we use the result of MRF summarized in \eqref{eq:MRF-Prob}. In practice, if the edge matrices are big to keep, one can store average (or a summary) of that matrix. Our experimental result shows the high performance of this method.

\section{Distributed Implementation}
\label{sc:distributed}
The stratified sampling provides the benefit of improved accuracy over the uniform sampling given the same sample size constraint. However the accuracy is still dependent on the sample size. In order to handle a larger sample size, we distribute the sample over a set of servers. As the queries are trivially parallelizable, the overall system latency is significantly reduced. We are also able to support many simultaneous queries over the same sample. We describe the requirements of the distributed system in \S~\ref{dist:requirements}. The system's architecture is introduced and explained in \S~\ref{dist:architecture}. \S~\ref{dist:analysis} contains a brief analysis on how the design fulfills the requirements.

\subsection{System requirements}
\label{dist:requirements}
The following non-functional requirements try to capture the most important necessities of the system:
\begin{enumerate}
	\item \textbf{Near real-time}: System response is expected to be in order of seconds.
	\item \textbf{Progressive result update}: The results are updated in realtime as the scanning of the sample progresses. The major advantage of this approach is that users are able to get a very quick estimate of the results before the full computation finishes. Our system generates partial query results in order to satisfy this requirement.
	\item \textbf{Fault tolerance}: In presence of network or hardware failures, the system should still produce results, at potentially reduced speed or accuracy.
	\item  \textbf{Scalability}: We consider two main aspects:
	\begin{enumerate}
		\item Horizontal scalability: Adding more servers should allow the system to execute on the sample faster or execute on larger samples for higher accuracy.
		\item Increased query handling: As the system usage increases due to users or automated software services, the system can handle the increased number of query requests without significant performance or accuracy degradation.
	\end{enumerate}
\end{enumerate}

There are additional requirements that describe availability and reliability scenarios, such as the guarantee that generating and distributing a new sample will not affect currently running reports. Describing these requirements and the design decisions concerning these are outside the scope of this paper.

\subsection{System architecture}
\label{dist:architecture}
The distributed system consists of two kinds of  nodes: \emph{counter nodes} and \emph{aggregate nodes}. The counter nodes are responsible for going over the part of the stratified sample present on that node and perform the necessary query. The aggregator nodes choose a set of counter nodes from a pool of counter nodes, distribute the query request to each selected counter node, receive partial results from counters and aggregate them into a single final result. This simple distinction of responsibilities allows the system to process increasing the amount of data with only adding a small constant overhead of network communication and the increased time of partial report aggregation as new servers are introduced. This approach has been evangelized by well known search engines \cite{barroso2003web} effectively.

\begin{figure}[t]
 	\centering
	\includegraphics[width=0.5\textwidth]{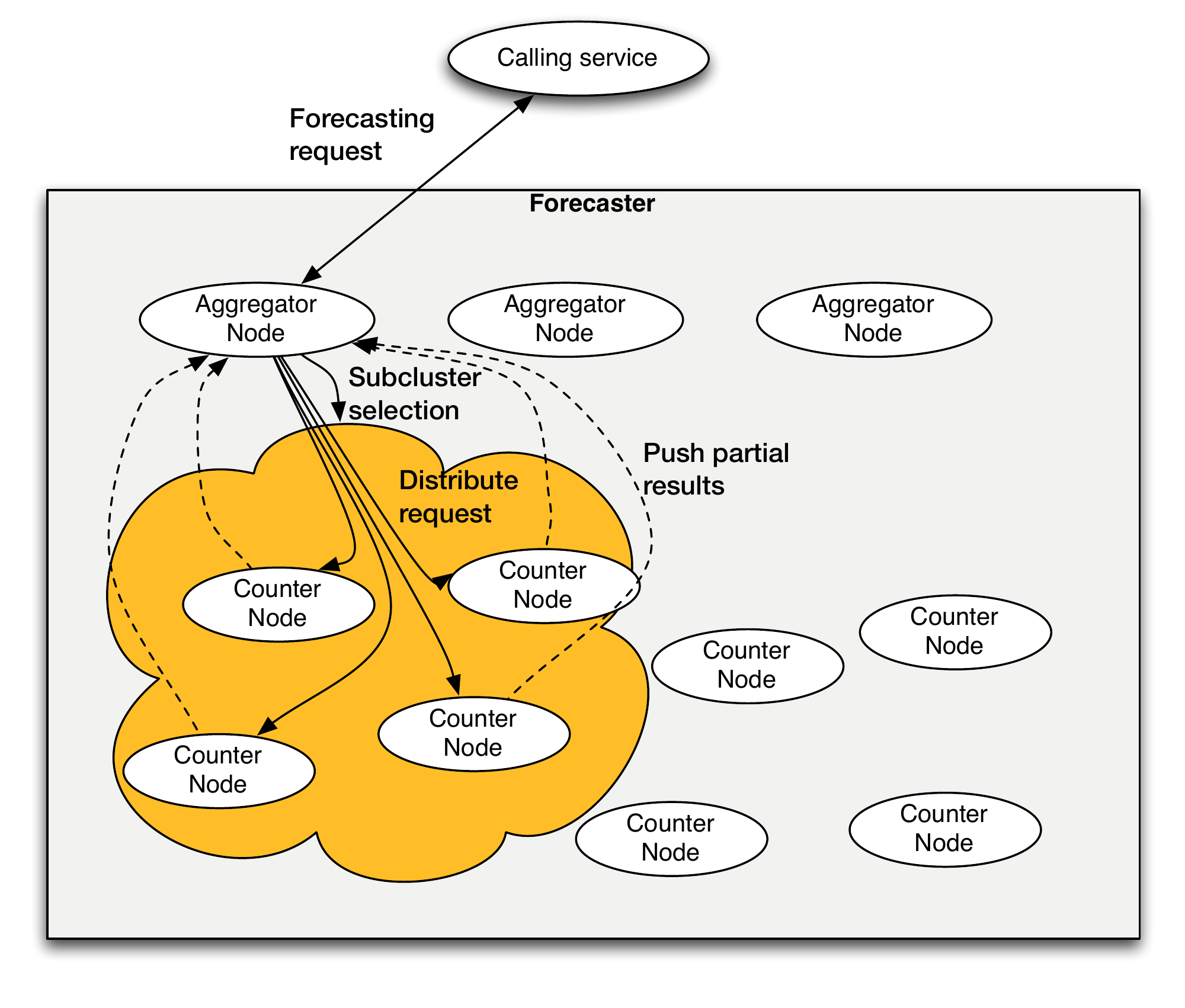}
	\caption{Initializing a report. The outside service calls an aggregator node to start the report. The aggregator node selects a sub-cluster of counter nodes and further distributes the request to them. This call starts a response thread on the counter machines that periodically send partial responses back to the aggregator. This ensures the aggregator has fresh data periodically, and the outside service can fetch the fresh results.}
	\label{call_flow}
\end{figure}

Figure~\ref{call_flow} shows the flow of the system. Other services can make requests to any of the aggregator nodes with all the targeting information used in the forecasting. The aggregator node will select a sub-cluster of counters and distribute the request. The counters will add these requests to a queue that is used to store and update all awaiting forecasting requests, see section \ref{dist:queue} for more details. The counters will also acquire a thread to push partial results frequently to the aggregator which distributed the request. Pushing the data instead of pulling on request reduces the amount of communication between the servers and can help in determining if a forecast prediction is already within a certain margin of error threshold. This support for early termination helps in reducing the overall system load on counter nodes, thus enabling higher query throughput.

\begin{figure}[t]
 	\centering
	\includegraphics[width=0.5\textwidth]{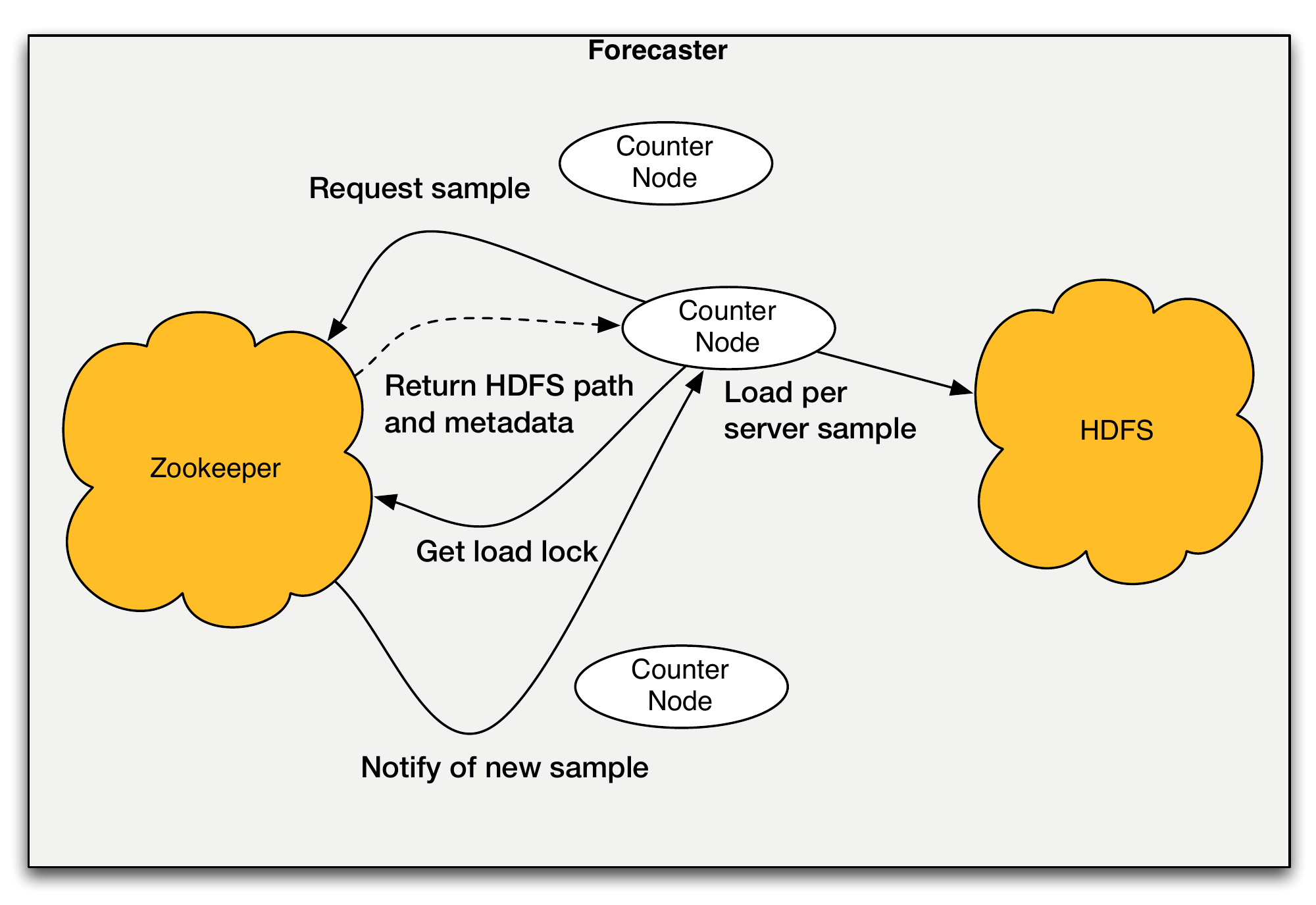}
	\caption{Managing the sample per server. A newly started node requests a sample from the Zookeeper cluster that returns the path to the sample acquired and some additional metadata. The node then loads the sample data and starts consuming it. Zookeeper will notify the node if a new sample is available. That will require the node to acquire a new sample. The node further requires permission to stop accepting report requests and start loading the data. This is done through a distributed semaphore through the Zookeeper cluster.}
	\label{sample_flow}
\end{figure}

Meanwhile, the stratified sample is generated periodically offline by a Map-Reduce process and stored on a Hadoop File System (HDFS) that can be accessed from all nodes. Distribution of the sample between the counter nodes and the notification when a new set of sample data is available is handled by a Zookeeper \cite{hunt2010zookeeper} cluster, a distributed lock service widely used in the industry. The sample itself is divided up into per server sub-samples, making sure the tuples in the per server sample is disjoint, and is further partitioned into smaller fragments. As each sub-sample has exactly similar strata present, the computation at each counter node can be performed independently and can be scaled up properly. When a counter node fails, the  sub-sample is temporarily lost from the whole system until a new counter node takes over, which can potentially reduce the accuracy the forecast in that time interval. The per server sample is loaded up in memory, since the IO overhead can be significant for systems dealing with large data sets. Zookeeper is used to orchestrate loading the updated sample data, ensuring that not all the counters are unavailable at the time of loading. This flow is shown in Fig.~\ref{sample_flow}.

\subsubsection{Sample management pipeline}
\label{dist:sample}
Managing the servers and the sample is done through Zookeeper in the following way:
\begin{enumerate}
	\item Counter nodes \textbf{acquire sub-samples} through a leader selection process controlled by the Zookeeper service. This prevents the sub-sample from being loaded in multiple counter nodes.  Having the sub-sample loaded in multiple counter nodes will lead to reduced overall accuracy. Given a fixed set of counter machines, it is always better to have each machine loading a separate sub-sample, as we handle any non-responding counter node gracefully. The Zookeeper service is also responsible for alerting when new offline sub-samples are generated.
	\item \textbf{Aggregator selection} is managed through a distributed semaphore, ensuring a fixed number of aggregator nodes among a pool of aggregator nodes to be present.
	\item Zookeeper also \textbf{orchestrates the loading} of the sample in a staggered fashion so that not all counetr nodes are non-responsive during the sub-sample reload time.
\end{enumerate}

Newly acquired sub-samples are loaded into counter node memory using multiple threads, both for performance reasons and to set up multiple threads for consuming the data. It is assumed that memory is scarce, at least scarce enough that two per server sample cannot be loaded into memory at the same time. This dictates our design decision of making sure that the counter node stops taking new requests from the aggregator, then waits for the existing requests to be finished and then only start reloading the new sub-sample in memory. This counter node is ready for new request only after the reload has happened successfully.

\subsubsection{Request queue pipeline}
\label{dist:queue}
Each forecasting request received by any of the forecasting nodes are wrapped in a collector and added to a queue. The collectors are responsible for gathering the set of partial results, whether it is done by processing the sample as in the case of counter nodes, or continuously receiving partial results from other nodes as in the case of aggregator nodes.

In case of counter nodes, the collector orchestrates a set of threads processing data that is fed by consumer threads iterating over the data. This strategy of consuming and processing data is useful in our use case, namely when processing data points require compressing and decompressing for memory optimization. This strategy further ensures that the system can scale with additional requests as, besides the extra overhead of evaluating the data points for the new request, there is no additional cost of optimizing the data. The data consuming and processing threads used are symmetric, meaning each consumer pushes data to a specific processing thread, which makes synchronization and locking unnecessary. Every time a partial result needs to be pushed, the results of the processing threads are merged and sent to the aggregator.

The aggregator collectors are responsible for keeping track of partial responses received for a request and making sure all information about the sample of the counters involved is at hand for accurate scaling of the results. Counter nodes push the partial results instead of a delta between pushes to ensure that losing a partial result during network communication is acceptable. This also means that the aggregator collector needs to merge all results from the counter nodes whenever a response is requested.

Collectors are considered to be finished when either (a) the full set of the sample has been processed, (b) an acceptable threshold is reached in terms of prediction error, or (c) the result was not requested for a longer amount of time. The queue, in case of every node, runs a maintenance thread on itself to check if any of the collectors are done. Those that are done are removed from the queue and put into a cache for some time so the results remain available before removing it completely from the system. These results are currently not persisted.

\subsubsection{Handling information on the sample for multiple servers}
\label{dist:stats}
It is not enough for the counter nodes to know about the sample they have acquired, it is also important that other nodes, especially aggregator nodes, know it too so that they can successfully scale the results. It might also be the case that a node did not load the full available sample into memory. This has the benefit to fine tune the system in terms of performance and accuracy. To ensure that the latest information is available, nodes automatically broadcast this to every other node. Whenever a node comes online it also both receives and sends this information to every other node.

On the other hand, ongoing forecasts should not be affected by these changes. Assuming a scenario where a counter finished producing a result, pushed it to the aggregator, and loaded up a new sample, the aggregator should still use the old sample information for scaling. For this reason every aggregator collector stores not only the partial results received for each collector, but also the summed sample information of all nodes participating in the forecast. 

\subsection{Analysis of the design}
\label{dist:analysis}
Section \ref{dist:requirements} detailed the requirements of the system. The analysis below describes how these requirements are fulfilled by the above described system and the various tradeoffs.

\begin{table*}[t]
\centering
\begin{tabular}{|l|l|}
\hline
\textbf{Message type} & \textbf{Description}\\
\hline
Initiate report & Message used to initiate a forecast report generation\\
Fetch result & Fetches partial or complete end result\\
Partial result & Message containing the partial results from a counter node\\
Sample information & Broadcast sample information\\
Distribute request & Distributes request to counter nodes\\
\hline
\end{tabular}
\caption{Message metrics}
\label{tab:messages}
\end{table*}

\subsubsection{Performance}
 The performance of the system can be modelled as the overall load on the system defined by
\begin{equation}
\begin{aligned}
t_d + t_s + t_c + t_m = \mathcal{O} (N + \frac{S}{N}),
\end{aligned}
\label{eq:dist_performance}
\end{equation}
where, $t_d$ is the time to distribute the request among $N$ counter nodes, $t_s$ is the time to scan the sample per node, $t_c$ is the communication latency overhead, $t_m$ is the time to merge the results on the aggregator node for a sample of size $S$. The above equation can be used to fine-tune the system. Lowering the number of querying a certain result increases the performance, but slows down the progressive report. Same is achieved by increasing the delay between the counter nodes reporting partial results. The greatest flexibility comes with tuning the sample size available per counter nodes, since $t_m$ is the most significant contribution to the overall system load. Greater representative sample may increase accuracy especially for more granular filtering, while small sample speeds up the report generation as counters finish faster. It is interesting to note that by adding more counters, $t_s$ for $N$ nodes is approximately the same as $t_s$ for $N\!+\!1$ nodes since the nodes work in parallel and the communication overhead is significantly less that the cost of processing the sample.

Since there are no database interactions and no disk IO operations, the system can be easily tuned to process a required size sample within seconds.

\subsubsection{User experience}
Forecasting can potentially take multiple tens of seconds so it is important to reduce the perceived waiting time for users. There are multiple research showing that having incremental page loading capability reduces the perceived waiting time significantly~\cite{bhatti2000integrating}. In our system we strive to implement incremental result publishing so that users can have a significantly good estimate of the query result very quickly. This is very important during the initial advertising campaign setup when the user is trying to explore several what-if scenarios.
Since aggregator nodes are continuously receiving partial results from counters that they can merge and scale using how much of the sample has been processed, partial results are available at any time.

\subsubsection{Availability}
There are multiple ways the design ensures fault tolerance. Through Zookeeper, we can monitor how many nodes are active and alive, and can distribute responsibilities accordingly. A node failure can result in a temporary decrease in performance or accuracy until a new node is introduced. In case of aggregator nodes, it is possible that some report generation fail as the node fails. This is deemed acceptable, as the failure happens quickly; the failed forecasting reports are easy to resubmit. As long as another aggregator node is available, the request can be served. 

It is possible that in extreme circumstances, too many nodes fail simultaneously. While the counter nodes are able to receive requests just by themselves and produce a fully usable result, the potential margin of error for just one machine might be too large for the report to be usable. However, we also provide a margin of error with all of our estimates and the user can choose to ignore the results if the error bounds are out of the expectation.

\subsubsection{Scalability}
Scalability is defined in two different ways in requirements. The ability to add servers with ease to forecast faster or with greater accuracy is achieved by the division of responsibilities. Since adding new counter nodes introduces only a small overhead in communication, the sample size can be increased for greater accuracy. It is also possible to redistribute the current sample to include the newly added node, which decreases the time needed to process the sample per server.

Handling increased amounts of forecast queries is achieved by considering two different factors: (1) Because new aggregators can be introduced to handle the outside communication and to make sure no request is lost, still providing the callers with the possibility of serving partial results, and (2) because the size of the sub-cluster used for forecasting can be tuned. It is possible to select a smaller sub-cluster of counter nodes so that the nodes are not exhausted, but potentially reducing the accuracy of the forecast.

\section{Experimental Results}
\label{sc:exp}
We provide two sets of experimental results. One set of results show the improvement of our sampling strategy over uniform and simple stratified sampling, while the other set of results is concerned with the distributed system implementation.

\begin{figure}[t]
 	\centering
	\includegraphics[width=3.2in]{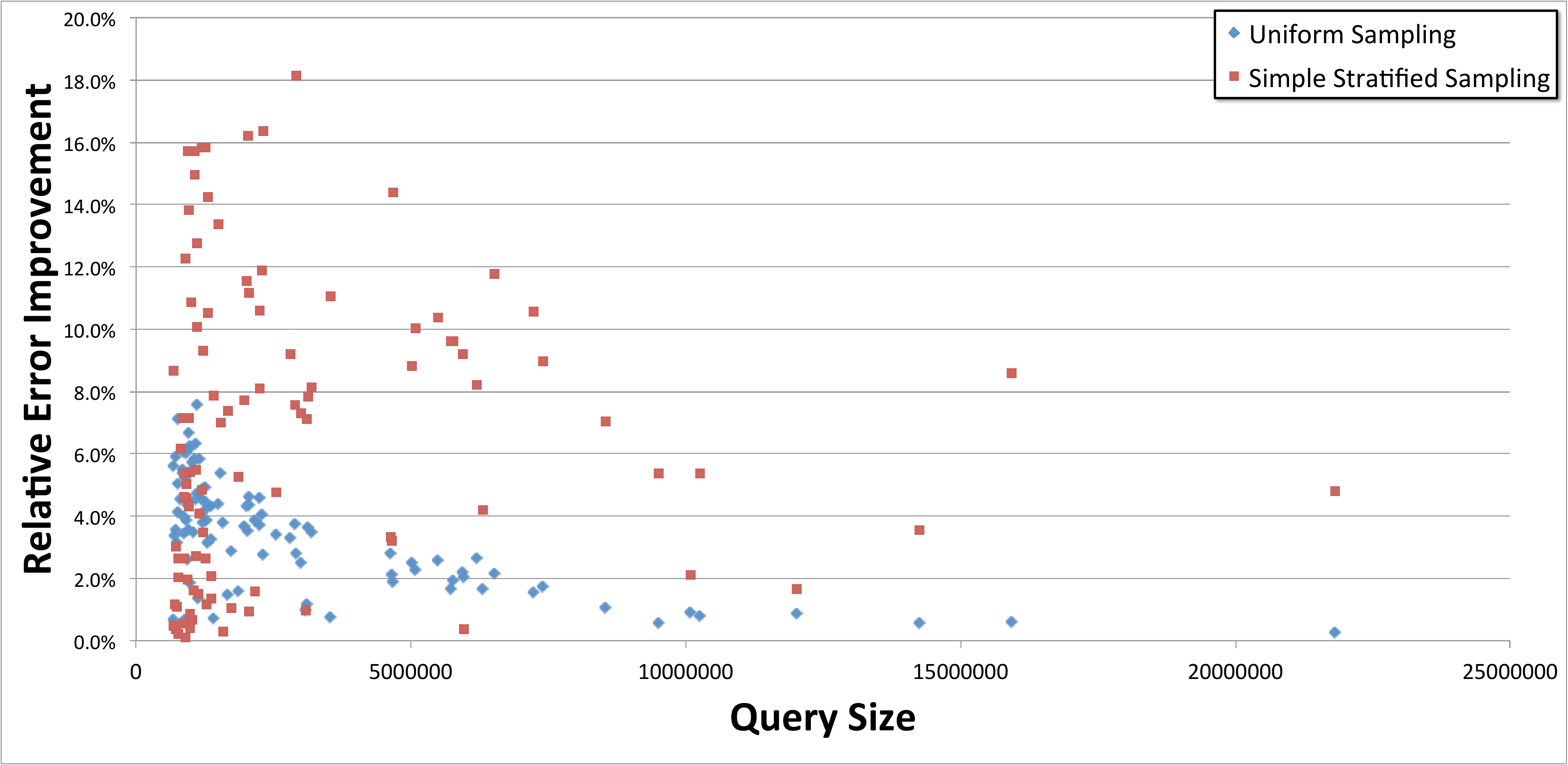}
	\caption{Illustration of the ratio of the error in estimation of queries on the sample generated by our proposed algorithms versus uniform and simple stratified sampling. Each point in the graph shows the ratio of the error of our sampling to either of methods with respect to the size of the query.}
	\label{fig:sampling-result}
\end{figure}

\subsection{Sampling Performance}
We illustrate the improvement of our sampling algorithm over uniform sampling and simple stratified sampling. The simple stratified sampling is our algorithm minus the fuzzy fall-back step. We plot the relative error of queries with respect to the size of the query. We make a sample of size $n=300K$ out of a total population of $N=1.5B$ feature vectors. Fig.~\ref{fig:sampling-result} shows that our sampling error is less than 8\% of the uniform sampling error and less than 18\% of the simple stratified sampling error on a query of size $1M$.

Obviously, the estimation error decreases as the size of the query increases but our relative (to other methods) error is also decreasing. This shows that our method not only performs better than the other two methods but also provides more advantages as the size of the query increases. The amount of our improvement is less comparing to the simple stratified scheme because the simple stratified scheme is using our reduced set of features for stratification.

\subsection{System Performance}
In Eq.~\eqref{eq:dist_performance} some metrics are introduced as the describing factors of system performance. We run experiments to get a sense of how the system behaves by generating reports with different filtering criteria.

Table~\ref{tab:messages} shows the messages communicated within the system. Since \emph{Initiate report}, \emph{Sample information} and \emph{Distribute request} messages are rare compared to \emph{Partial result} and potentially \emph{Fetch response}, it is more important to focus on the performance of the latter two. Our measurements show that fetching the results take a significantly more time than pushing the partial results from counter nodes to the aggregator nodes. This is not due to the message size that is almost identical to the size of partial results but is due to the time it takes to merge the results.

\begin{figure}[t]
\centering
\includegraphics[width=0.5\textwidth]{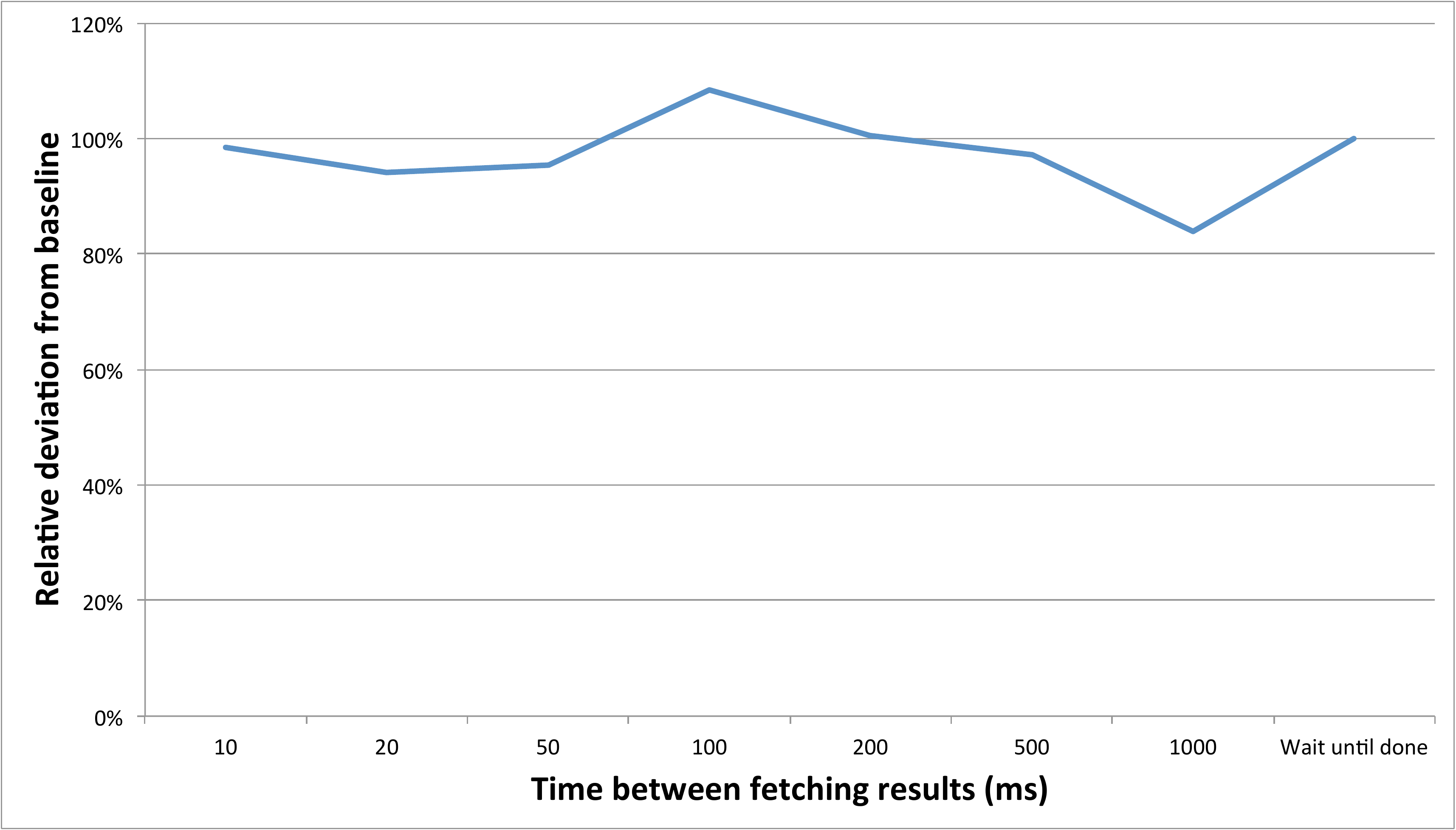}
\caption{Influence of fetching results more frequently on the system performance}
\label{tab:fetch_time}
\end{figure}

In trying to see how much influence fetching results have on the overall system, Table~\ref{tab:fetch_time} shows how the total report generation time changes depending on the interval the results are requests. The table shows the relative performance deviation from the baseline case where no fetching is done. According to the results, there is no significant impact on the overall performance of the system due to more frequent merging of partial results.

Out of the terms defined in Eq.~\eqref{eq:dist_performance}, $t_s$ has the greatest impact on the total runtime. Comparing the distributed system with three counter nodes against a simple machine solution that consumes the same sample without any communication overhead showed about 33\% increase on average. Improving the performance and the parallelization is a topic for future work.

\section{Conclusions}
\label{sc:conclusion}
Forecasting is key for advertisers to pinpoint their exact targeting constraints and budget spend so as to reach the best segments of the audience with the optimum valuation. As such, forecasting has to be accurate. Moreover, forecasting needs to be fast to allow advertisers to perform interactive exploration while finalizing their campaign parameters during campaign setup. Forecasting is also useful to evaluate the DSP partner of the advertiser to compare a before and after picture of what was forecasted and how the campaign returned. The same evaluation is also used by DSPs during debugging campaign delivery and performance issues.

In this paper, we propose a forecasting solution to solve for both accuracy and speed. Our Markov Random Fields based algorithm perform an accurate stratified sampling of the audience seen by the DSP. We offer a fuzzy fall-back mechanism to make the algorithm more practical for small-sized strata. Our distributed solution enables us to run forecasting queries in seconds and to tolerate faults with servers in the distributed cluster. Experimental results from the deployed system validate the efficiency and efficacy of the proposed solution.

Future work includes adding support for ad hoc querying the forecasting sample.

%ACKNOWLEDGMENTS are optional
\section*{Acknowledgments}
The authors would like to thank Changgull Song for testing the entire framework in the staging environment and Kuang-Chih Lee for his valuable comments.

\bibliographystyle{natbib} 
\bibliography{kdd2013}
\end{document}